\documentclass[10pt,twocolumn,letterpaper]{article}
\usepackage{cvpr}
\usepackage{times}
\usepackage{epsfig}
\usepackage{graphicx}
\usepackage{amsmath}
\usepackage{amssymb}

\usepackage[pagebackref=true,breaklinks=true,letterpaper=true,colorlinks,bookmarks=false]{hyperref}

\cvprfinalcopy % *** Uncomment this line for the final submission
\def\cvprPaperID{3929} % *** Enter the CVPR Paper ID here

\ifcvprfinal\fi
\begin{document}

%%%%%%%%% TITLE
\title{A Mutual Learning Method for Salient Object Detection with intertwined Multi-Supervision}

\author{
Runmin Wu,
Mengyang Feng,
Wenlong Guan,
Dong Wang,
Huchuan Lu,
Errui Ding\\
\textsuperscript{1}Dalian University of Technology\\
\textsuperscript{2}Department of Computer Vision Technology (VIS), Baidu Inc.\\
\small\texttt{josephinerabbit@mail.dlut.edu.cn},\
\small\texttt{wdice@dlut.edu.cn, lhchuan@dlut.edu.cn, dingerrui@baidu.com}
}

\maketitle
%\thispagestyle{empty}

%%%%%%%%% ABSTRACT
\begin{abstract}
Though deep learning techniques have made great progress in salient object detection recently, the predicted saliency maps still suffer from
incomplete predictions due to the internal complexity of objects and inaccurate boundaries caused by strides in convolution and pooling operations. To alleviate these issues, we propose to train saliency detection networks by exploiting the supervision from not only salient object detection, but also foreground contour detection and edge detection. First, we leverage salient object detection and foreground contour detection tasks in an intertwined manner to generate saliency maps with uniform highlight. Second, the foreground contour and edge detection tasks guide each other simultaneously, thereby leading to precise foreground contour prediction and reducing the local noises for edge prediction. In addition, we develop a novel mutual learning module (MLM) which serves as the building block of our method. Each MLM consists of multiple network branches trained in a mutual learning manner, which improves the performance by a large margin. Extensive experiments on seven challenging datasets demonstrate that the proposed method has delivered state-of-the-art results in both salient object detection and edge detection. 
\end{abstract}

%%%%%%%%% BODY TEXT
\section{Introduction}

Salient object detection aims to segment the most distinctive object regions in a given image, acting as an important pre-processing step for various vision tasks, including image captioning~\cite{Fang2015From},  visual tracking~\cite{Borji2012Adaptive}, visual question answering~\cite{vqa}, and person re-identification~\cite{reid}.

By integrating multi-scale features, previous deep learning based methods~\cite{Zhang2017Amulet,DSS,DHS,RFCN} are capable of detecting the salient objects from a global and coarse view.  Though good performance has been achieved, they still suffer from two major drawbacks:
1) the entire salient objects can hardly be uniformly highlighted due to their complex internal structures; 
2) the prediction around object contours are inaccurate due to the information loss caused by strided convolution and pooling operations.

\begin{figure}[t]
		\begin{tabular}{c@{}c@{}c@{}c@{}c@{}}
			\includegraphics[width=0.18\linewidth, height=0.3\linewidth]
			{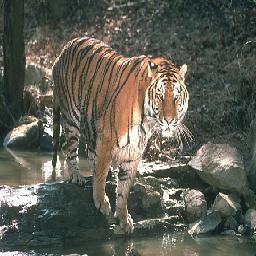} &									  					    \includegraphics[width=0.18\linewidth, height=0.3\linewidth]            		{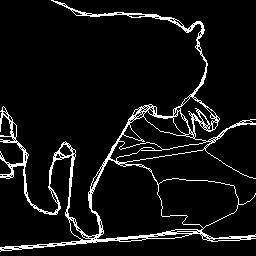} &
			\includegraphics[width=0.18\linewidth, height=0.3\linewidth]
			{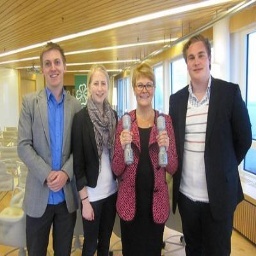} & 
			\includegraphics[width=0.18\linewidth, height=0.3\linewidth]
			{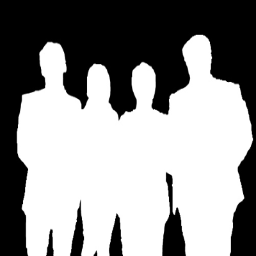} & 
            \includegraphics[width=0.18\linewidth, height=0.3\linewidth]
			{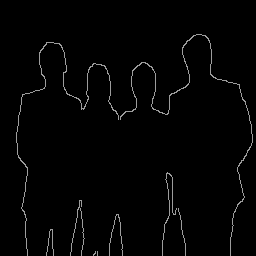} \\
			{\footnotesize (a) E} & 
			{\footnotesize (b) E-gt} & 
			{\footnotesize (c) S} & 
            {\footnotesize (d) S-gt} & 
            {\footnotesize (e) FC-gt}\\
		\end{tabular}
		\centering
		\caption{The examples of our two inputs and three supervisions. E and E-gt are collected from edge detection dataset. S and S-gt are collected from salient object detection dataset, and FC-gt is the foreground contour extracted from S-gt.} 
\label{fig:fig2}
%\vspace{-2mm}
\end{figure} 

To address the above-mentioned issues, we present a new training strategy and a network design which simultaneously leverage three tasks, including salient object detection, foreground contour detection and edge detection. We feed two images into our network, one for salient object detection and another for edge detection, with three kinds of corresponding supervisions. The two input images are collected from salient object detection and edge detection datasets respectively. For description convenience, we use E and E-gt to denote the input image and ground truth from edge detection dataset, and S and S-gt for the input image and ground truth from salient object detection. Additionally, the foreground contour ground truth, denoted as FC-gt, is extracted by Canny~\cite{Canny} operator on the salient object masks (S-gt). The two input images and three supervisions are shown in Figure ~\ref{fig:fig2}.

First, to generate the saliency maps entirely with uniform highlight, we employ salient object detection and foreground contour detection in an intertwined manner. 
These two tasks are highly correlated since both demanding accurate foreground detection. Nevertheless, they are also different from each other in the scenes that saliency detection involves dense labeling (i.e., ``filling'' the internal of object regions) and is more likely to be affected by the internal complexity of salient objects, giving rise to uneven foreground highlight. In contrast, the foreground contour can be ``extracted'' based on low-level cues, such as edges and textures, given the rough location of foreground objects. 
Therefore, the contour detection task is more robust to objects' internal structures, but may be misled by rich edge informations around object contours.

In the proposed intertwined strategy, the two tasks are interlaced at different blocks of the network, forcing the network to learn ``filling'' and ``extracting'' the foreground contour alternatively. As a consequence, the network can benefit from the strengths of both tasks and overcome their defects, leading to more entirely highlighted salient regions.

Second, to alleviate the blur boundary issue in the predicted saliency maps, we propose to improve foreground contour detection with auxiliary supervision from edge detection task. To this end, an edge module is designed and jointly trained with our backbone network. Since the edge module takes as input the saliency features from our backbone network, the semantic information encoded in these features can effectively suppress noisy local edges. Meanwhile, the edge features extracted by the edge module serve as additional input to foreground contour detection, ensuring more accurate detection results with low-level cues.

Third, to further improve performance, we propose a novel Mutual Learning Module (MLM) inspired by the success of Deep Mutual Learning (DML)~\cite{DML}. A MLM is built on top of each block of our backbone network and comprises of multiple subnetworks which are trained in the peer-teaching strategy by a mimicry loss, yielding an additional performance gain.

Our main contributions can be summarized as follows.
\begin{itemize}
	\item We propose to train deep networks for saliency detection using multi-task intertwined supervision, where foreground contour detection is effectively leveraged to render accurate saliency detection.  
    \item We employ the foreground contour detection and edge detection task to guide each other to generate more accurate foreground contour and reduce the noise in edge detection simultaneously.
	\item We design a novel network architecture named Mutual Learning Module, which can better leverage the correlation of multiple tasks and significantly improves saliency detection accuracy.    
\end{itemize}
We compare our method with 15 state-of-the-art salient object detection on 7 challenging datasets and 6 popular edge detection methods on BSD500\cite{ucm}. The results show that the proposed algorithm performs much better than other competing saliency models, meanwhile, it achieves comparable edge detection performance with much faster speed.
\begin{figure*}[ht]
\centering
\includegraphics[width=\textwidth, height=0.6\linewidth]{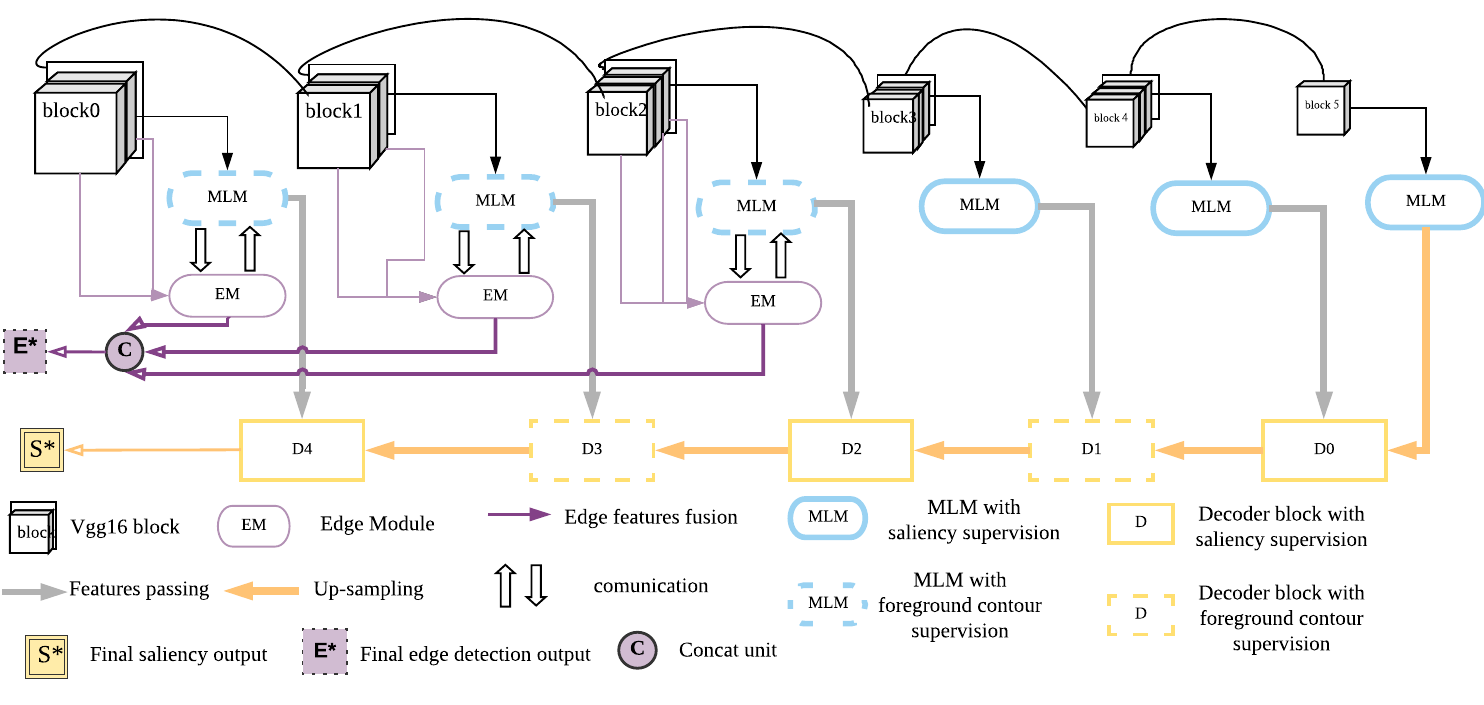}\\
        \vspace{8mm}
		\centering
		\caption{The overall framework is consist of the VGG-16 (block0$\sim$block5), 6 mutual learning modules (MLMs), 3 edge modules (EMs) and a decoder (D0-D4). We employ deep supervision at each MLM, EM and decoder block. Here, the module with dashed frame denotes the block supervised by FC-gt and the module with the solid frame supervised by S-gt or E-gt. In our revised version, all the decoder block have mask supervision meantime the first two decoder block are supervised with FC-gt additionally. For the encoder we use the interwined way for supervision.} 		
        
\label{fig:Net}
     %   \vspace{-1mm}
\end{figure*} 
\section{Related Work}
Effectively utilizing spatial details and semantic information is a crucial factor in achieving the state-of-the-art performance on salient object detection. 
Most existing methods use \textit{skip-connection} or \textit{recurrent architecture} to integrate hierarchical features from convolution Neural Networks (CNNs). They can roughly detect the targets but can not uniformly highlight the entire objects, and also suffer from the blur boundary. To generate clear boundaries, some methods attempt to introduce extra edge information to the saliency network.
\subsection{Integrating Hierarchical Features}
\subsubsection{Skip-connection}
%Skip-connection is first proposed in HED~\cite{hed} for edge detection.
In HED~\cite{HED}, the authors propose to build \textit{skip-connections} to exploit multi-scale deep features for edge detection. The edge detection is an easier task since it does not rely too much on high-level information. On the contrary, the salient object detection is in great need of semantic features. Thus directly introducing \textit{skip-connections} into salient object detection is unsatisfactory.
A promoted version called \textit{short-connections}, proposed in DSS~\cite{DSS}, resolves this issue by linking the deeper layers towards shallower ones and skipping the middle ones.  Another work by SRM~\cite{SRM} proposes a stage-wise refinement model and a pyramid pooling module to integrate both local and global context information for saliency prediction.
In this way, the multi-scale feature maps can assist to locate salient targets and recover local details  more effectively.
%In particular, the stage-wise model is utilized to add lower level detailed features to the predicted map stage by stage. 
\subsubsection{Recurrent Architecture}
A Recurrent Fully convolution Network (RFCN)~\cite{RFCN} is proposed to incorporate saliency prior knowledge for more accurate inference. It can refine the saliency map by remedying its previous errors through the recurrent architecture. A recurrent convolution layer (RCL) is introduced in DHS~\cite{DHS} to progressively recover image details of saliency maps through integrating local context information. 

Most of these methods attempt to find a way to fuse multi-scale features for better recognizing salient and non-salient regions. However, supervised by S-gt only, they still fail to generate a binary map as the ground truth. Because that the pixels inside the saliency region could be complex due to various cues, such as illumination, color etc.. The lower level features are more susceptible by these factors, which makes each pixel has different responses at different scale. As the results, the fusion strategy could not capture the entire object. In this paper, we explore to learn the saliency detection and foreground contour in an intertwined manner, which enables our network to recognize the overall shape of the salient targets and uniformly assign foreground labels to the entire objects.
\subsection{Exploiting Edge Information}
There are some recent attempts exploiting extra edge information for saliency detection to generate prediction maps with clear boundary.
In \cite{edgeaware}, the authors make a new dense label of three categories(saliency objects, saliency objects' boundaries and background) to emphasis the accuracy of saliency boundary detection. 
Moreover, they introduce extra hand-craft edge features as a complementary to preserve edge information effectively. 
In~\cite{edgeregion} and~\cite{edgeregion2}, edge knowledge is used for detecting region proposals. They first exploit a pre-trained edge detection model to detect objects' edge. Then based on the detected edge, they segment the input image into edge regions (similar to super-pixels), and generate saliency score map in every region via a mask-based Fast R-CNN. With the edge-region based method, the models succeed to preserve the objects' boundary in saliency detection.

However, all these existing edge-based methods only utilize prior edge knowledge to help saliency task. In contrast, our model utilizes the edge detection  and foreground contour detection to guide each other simultaneously, which demonstrates that the foreground contour detection also benefits the edge detection. As the results, our method achieves comparable performance while running faster than other edge detection methods.
\section{The Proposed Method}
\subsection{Architecture Overview}
 %In this section, we elaborate the proposed framework. We first describe our alternative supervision strategy with Sal-gt and SalB-gt. Then, we give detailed depictions of our mutual learning and joint learning methods. Finally, we describe our feature aggregation method, a decoder network.
Our network follows the encoder-decoder architecture and consists of four main components, which are  VGG~\cite{simonyan2014very} backbone, mutual learning module (MLM), edge module (EM) and the decoder blocks. The overview of our network is shown in Figure~\ref{fig:Net}.

The encoder part consists of a VGG-16 backbone, six mutual learning modules and three edge modules.
For the VGG-16 backbone, we discard the layers after $\mathtt{pool5}$ and denote the remaining blocks ($\mathtt{conv1}\sim\mathtt{conv5}$, $\mathtt{pool5}$) as block0$\sim$block5. We build six MLMs on top of each blocks to extract the foreground contour features and saliency features. Additionally, we apply an EM at every block to extract edge features, and each EM is connected with all convolution layers in corresponding VGG block. We use a residual architecture to transfer features between the EM and the corresponding MLM. 

As for the decoder, we employ a deeply supervised framework similar to the U-net for fusing multi-scale features from MLMs and generating predictions. The five decoder blocks (D0$\sim$D4) share the similar structure. Each of the decoder block aims to fuse the features from MLM and the upsampled features from the previous block. It merges the input features and outputs upsampled features to the next block by a deconvolution layer. In addition, for deep supervision, we employ a convolution layer to produce a predicted map at each block.
%, and the features contacted are also delivered to the next block as input by a deconvolution layer. 
%Additionally, each of decoder blocks has one convolution layer to generate saliency predictions (object mask or boundary map) denoted as $D^{i}$ , where $i$ indexes the decoder blocks. Sal-gt and SalB-gt are used as the alternative supervisions. The detail settings can be found in our supplementary material.
\subsubsection{Mutual Learning Module}
The Mutual Learning Module aims at improving the performance of saliency detection and foreground contour detection inspired by the success of Deep Mutual Learning~\cite{DML}. 
The method DML sets several student networks for a same task and utilizes their predictions as sub-supervisions for each other. Each student network is a complete model trained in a mutual manner, but can work independently.

In contrast, each student network in our Mutual Learning Module (MLM) is a simple sub-network consisted of three consecutive layers, which is aimed at extracting features and generating a prediction, as the Figure~\ref{fig:ML} illustrated. The detailed settings are shown in supplementary material. We set larger convolution kernels with dilation for the deeper MLMs to capture fine global information. The L2 distance is employed for mimicry loss in our model. 
\begin{figure}[t]
\includegraphics[width=0.6\textwidth, height=0.8\linewidth]{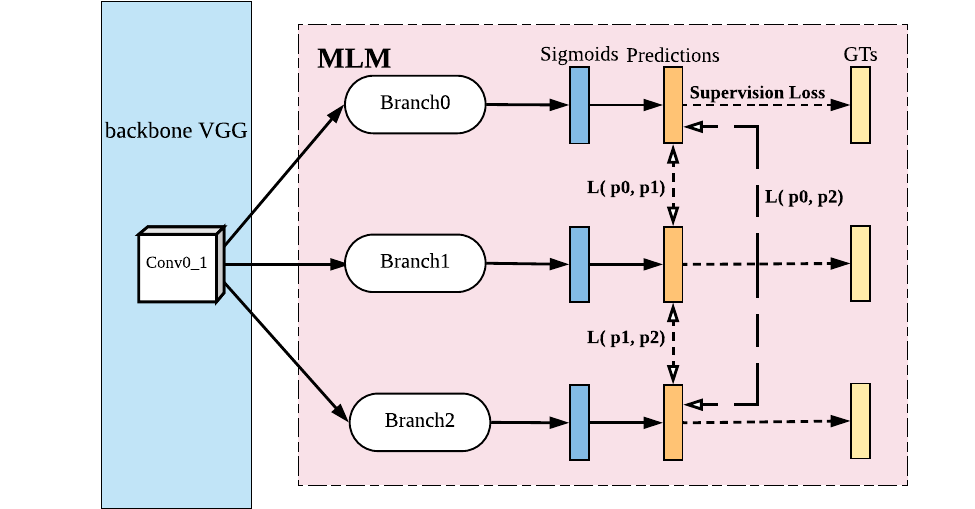}\\
        \vspace{8mm}
		\centering
		\caption{The example of Mutual Learning Module. Each student branch is supervised by ground truths and a L2-based mimicry loss used to match with every other branch. }

\label{fig:ML}
%\vspace{-4mm}
\end{figure}

Let $\upsilon_{i_j}$ denote the features in $j$th convolution layer of the block$i$, and $\phi_{s}^{i_{k}}$ represent the function of the $k$th student branch. The prediction $A_{s}^{i_{k}}$ can be given as:
\begin{equation}
\left\{
             \begin{array}{lr}
             A_{s}^{i_{k}}=Sigmoid(\phi_{s}^{i_{k}}(\upsilon_{i_1})), &  for \quad  i =0,1\\
             A_{s}^{i_{k}}=Sigmoid(\phi_{s}^{i_{k}}(\upsilon_{i_2})),& for\quad  i =2,3,4\\ \end{array}
\right.
\end{equation}
where $Sigmoid$ is the activate function, and the 0-based indexed mode is applied.

By learning from each other, MLM provides a softer loss for each sub-network, which leads to the parameters of each student networks converging to a better local minimal. Notice that several student branch networks are trained in a MLM, but we only use one of those student branches for testing by random. 

% Concretely, we apply the shallower three MLMs for foreground contour detection with the supervision of FC-gt based on the observation in ~\cite{show}. The deeper three MLMs are supervised by S-gt for saliency detection. The predictions in every MLM are shown in Figure~\ref{fig:side}.
% \begin{figure}[t]
% 		\begin{tabular}{c@{}c@{}c@{}c@{}c@{}c@{}c@{}c@{}}
% \includegraphics[width=0.12\linewidth, height=0.25\linewidth]
% 			{figures//fig2_1.pdf} &\includegraphics[width=0.12\linewidth, height=0.25\linewidth]
% 			{figures//fig2_2.pdf} &			\includegraphics[width=0.13\linewidth, height=0.25\linewidth]
% 			{figures//fig2_3.pdf} &
% 			\includegraphics[width=0.12\linewidth, height=0.25\linewidth]
% 			{figures//fig2_4.pdf} &
% 			\includegraphics[width=0.12\linewidth, height=0.25\linewidth]
% 			{figures//fig2_5.pdf} &
%             \includegraphics[width=0.13\linewidth, height=0.25\linewidth]
% 			{figures//fig2_6.pdf} &
%              \includegraphics[width=0.13\linewidth, height=0.25\linewidth]
% 			{figures//fig2_7.pdf} \\
% 			{\footnotesize S-E} &
% 			{\footnotesize block0} &
% 			{\footnotesize block1} &
% 			{\footnotesize blcok2} &
%             {\footnotesize block3} &
%             {\footnotesize block4} &
%             {\footnotesize block5 } \\
% 		\end{tabular}
% 		\centering
% 		\caption{The S-E (left) is the saliency boundary labels extracted by Canny from existing saliency segmentation ground truth. The block0 to block5 show the saliency outputs from corresponding 
% side out branches.}
% 		\label{fig:side}
% %\vspace{-2mm}
% \end{figure} 

\subsubsection{Edge Module}
The authors visualize neurons' receptive fields in a pre-trained VGG-19 model before and after fine-tuning for saliency detection task. The result shows 
that the responses of neurons at first three pooling layers hardly change after fine-tuning for saliency task. What's more, even fine-tuning the network for saliency detection, neurons' responses at each of the first three convolution layers still keep lots of similarities with edge patterns. It indicates that the first three blocks in the pre-trained VGG are suitable to capture both edge information and saliency information at the same time.
Based on this, we employ an extra edge module (EM) on each first three VGG blocks  for edge detection task and helping foreground contour detection in MLM.

Every two convolution layer in one block is connected to another convolution layer for extracting rich edge features, and features from different layers are merged to generate an edge probability map.

Specifically, for the input image E, every EM produces an edge probability map $A_{e}^{i}$, and the outputs are collected for merging into final edge prediction $E^*$. For the input image S, the EM only provides edge feature maps $a_{e}^{i}$ for the MLM. The two modules are connected in a residual manner, which is designed to reduce the noise in edge features for foreground contour detection.

For the $i$th block, denoting the function of EM as $\psi^{i}$, the output feature map $a_{e}^{i}$ and edge probability map $A_{e}^{i}$ can be given by:
\begin{equation}
\left\{
             \begin{array}{lr} 
             a_{e}^{i}=\psi^{i}(\upsilon_{i_0},\upsilon_{i_1}), & for \quad  i =0,1\\
             a_{e}^{i}=\psi^{i}(\upsilon_{i_0},\upsilon_{i_1},\upsilon_{i_2}),&  for \quad  i =2 \\
A_{e}^{i}=Sigmoid(a_{e}^{i}) &  \\
 \end{array}
\right.
\end{equation}

Receiving the information from EM, the MLM is capable to generate precise contour. In the meanwhile, the shared basic blocks provide semantic information from foreground contour detection and thereby helping the edge detection to ignore the useless local edges. We show more validations in Sec.~\ref{AbStd}. 
\subsection{Intertwined Supervision Method}
In previous MLMs, we employ the FC-gt as supervision at the shallower three MLMs for keeping fine detailed contour information and the S-gt as supervision at the deeper three modules to focus on semantic information.

We further apply two tasks alternatively at different decoder blocks in an intertwined manner. To be more specific, we set S-gt for supervision at D0, D2, D4, and FC-gt at D1, D3. Each decoder block is aimed at fusing the features from the previous block and corresponding MLM, and then transmitting the features to the next block. 
%Therefore, there are three kinds of staggered ways:
%(a) take in both features of salient object region from MLM and deeper block, and transmit them to the features of foreground contour;
%(b) take in the features of foreground contour from MLM and another features of salient object region from deeper block, and transmit them to features of foreground contour; 
%(c) take in both features of foreground contour from MLM and deeper block, and transmit them to features of saliency object region.

As illustrated in Figure~\ref{fig:Net},

%By utilizing FC-gt and S-gt supervision in this intermined manner, our method succeed to generate salient object predictions with uniform highlight and keep fine foreground contour. 
with the supervisions of FC-gt and S-gt in this intertwined manner, our method succeeds to generate saliency predictions with entire uniform highlight and keep fine foreground contours simultaneously.  

\subsection{Training Strategy}
\subsubsection{Loss Function}
In the deep supervision strategy, we resize the three kinds of ground truths for adapting the resolution of each stage. For simplicity, we treat salient object detection and foreground contour detection as saliency tasks and denote S-gt along with FC-gt as $\hat{S}$ and E-gt as $\hat{E}$. Our total loss is consisted of two parts, \textit{i.e.} the loss $\mathcal{L}_{Enc}$ in encoder and the one $\mathcal{L}_{Dec}$ in decoder.

Mathematically, the $\mathcal{L}_{Enc}$ can be given as,
\begin{equation}                   \mathcal{L}_{Enc}=\theta_{s}\mathcal{L}_{S}+\theta_{e}\mathcal{L}_{E}+\theta_{m}\mathcal{L}_{mimicry},
\end{equation}
where the $\mathcal{L}_{S}$, $\mathcal{L}_{E}$, and $\mathcal{L}_{mimicry}$ are the loss functions for saliency tasks, edge detection and the mimicry loss in MLMs, respectively, and the $\theta$s are their weights, which are set as 0.7, 0.2, 0.1.

Specifically, we employ Binary Cross-Entropy loss $l_{bce}$ for saliency tasks and edge detection. They can be wrote as, 
\begin{equation}          
\mathcal{L}_{S}=\sum_{i=0}^{5} r_{s}^{i} l_{bce}(A_{s}^{i},\hat{S}),
\end{equation}
\begin{equation}          
\mathcal{L}_{E}=\sum_{i=0}^{2} r_{e}^{i} l_{bce}(A_{e}^{i},\hat{E})+ l_{bce}(E^*,\hat{E}),
\end{equation}
where $r_{s}^{i}$ and $r_{e}^{i}$ represent the weights at $i$th block. 
%In our experiments,  the $r_{s}^{i}$ are simply set as 0.2 except the the $r_{s}^{5}$ which is set as 0.5.
For the mimicry loss,
we employ the MSE loss function $l_{mse}(a,b)=(a-b)^2$. Considering the six MLMs with $k$ student branches, the total mimicry loss can be given as,
\begin{equation}
\mathcal{L}_{mimicry} = \frac{1}{2}\sum_{i=0}^5\sum_{n=0}^{K}\sum_{m=0}^{K}r_{mlm}^{i}l_{mse}(A_{s}^{i_{n}},A_{s}^{i_{m}})
\end{equation}
where $r^i_{mlm}$ denotes weight at the $i$th block and $K$ denotes the number of the student branches.

As for the decoder, the weight at the $i$th decoder block is denoted as $r_{dec}^{i}$. Then $\mathcal{L}_{Dec}$ can be given as,

\begin{equation}          
\mathcal{L}_{Dec}=\sum_{i=0}^{4}r_{dec}^{i}l_{bce}(D^{i},\hat{S})
\end{equation}.

The detail settings of all above $r$s can be found in our supplementary materials.

\section{Experiments}

\begin{table*}[ht]
\vspace{8mm}
\centering
\caption{Quantitative evaluations. The best three scores are shown in {\color{red}{red}}, {\color{green}{green}} and {\textbf{black}}, respectively.}\label{tab:fmeasure_mae}
	\small
	\renewcommand{\arraystretch}{1.2}
	\begin{tabular}{|c|c|c|c|c|c|c|c|c|c|c|c|c|c|c|c|}
		\hline
		\ & \multicolumn{3}{c|}{DUTS}  & \multicolumn{3}{c|}{ECSSD} &\multicolumn{3}{c|}{SOD} &\multicolumn{3}{c|}{HKUIS}&\multicolumn{3}{c|}{OMRON}\\
		\cline{2-16}
		& $F_\beta$ & MAE & $S_m$ & $F_\beta$ & MAE & $S_m$ & $F_\beta$ & MAE & $S_m$ & $F_\beta$ & MAE & $S_m$& $F_\beta$ & MAE & $S_m$\\
		\hline

 %fb,mae,sm 
 %ECSSD     mae :  0.037010927 E-measure :  0.9444958636613956 S-measure :  0.903340654456433 Wgt-F :  0.9007830159342797 Adp-F :  0.9161483203562129
 
%DUT-test  mae :  0.04404416 E-measure :  0.9050440272765882 S-measure :  0.8525624353569886 Wgt-F :  0.8035924893910211 Adp-F :  0.8232688350197555

%SOD mae :  0.11024639 E-measure :  0.8060584683485804 S-measure :  0.7667445151671889 Wgt-F :  0.7394113352512389 Adp-F :  0.7942230068140005
%OMRON     mae :  0.060021527 E-measure :  0.8498925860708846 S-measure :  0.8070597859045221 Wgt-F :  0.7218584585459576 Adp-F :  0.7446085584291255

%hkuis mae :  0.030396774 S-measure :  0.899689477147407 Adp-F :  0.9058344511202915
Ours
&{\color{red}{0.825}}&{\color{red}{0.042}}&\bf{0.849}
%&{\color{red}{0.838}}&{\color{red}{0.069}}&\color{green}{0.849}&
&\color{red}{0.921}&\color{red}{0.037}&\bf{0.904}&
0.775&0.114&0.754
&\color{green}{0.904}&\color{red}{0.032}&0.894&
\color{green}{0.737}&\color{red}{0.057}&{0.800}\\

% Ours
% &{\color{red}{0.815}}&{\color{red}{0.046}}&\bf{0.849}
% %&{\color{red}{0.838}}&{\color{red}{0.069}}&\color{green}{0.849}&
% &\color{red}{0.917}&\color{red}{0.036}&\bf{0.906}&
% \color{red}{0.803}&\color{green}{0.104}&\bf{0.772}
% &\color{green}{0.904}&\color{red}{0.031}&\bf{0.900}&
% \color{green}{0.749}&\color{red}{0.059}&{0.810}\\

         BMPM	
         &0.751& \bf{{0.049}}&{\color{red}0.861}
        % &0.769&{{0.074}}&{\bf{0.845}}&
        & 0.869&{\bf{0.045}}&\color{green}{0.911}&
         0.763&0.107&\color{green}{0.787}&
         0.871&{{0.039}}&{0.907}&
         0.692&0.064&{0.809}\\
         
     	DGRL&0.768&0.050& 0.841
     %	&{\color{green}{0.825}}&0.072&{0.836}
     	&\bf{0.903}&\color{green}{0.041}&{0.903}
     	&\bf{0.799}&\color{green}{0.104}&{0.771}
     	&\bf{0.890}&\color{green}{0.036}&{0.895}
     	&\bf{0.733}&0.062&{0.806}\\
     	
        PAGR&\bf{0.788}&{0.056}&{0.837}&
      %  {0.807}&{0.093}&{0.818}&
        {0.894}&{0.061}&{0.889}
        &{-}&{-}&{-}
      &  {0.886}&{0.048}&{0.887}
        &{0.711}&{0.071}&{0.805}\\
        
        RAS&{0.755}&{0.060}&{0.839}
        %&{0.785}&{0.104}&{0.795}
        &{0.889}&{0.056}&{0.893}
        &\bf{0.799}&{0.124}&{0.764}
        &{0.871}&{0.045}&{0.887}
        &{0.713}&{0.062}&\color{green}{0.814}\\
        
        PiCANet&{0.755}&{0.054}&{\color{red}0.861}
       % &{0.801}&{0.077}&{\color{red}0.850}&
        &{0.884}&{0.047}&\color{red}{0.914}
        &{0.791}&\color{red}{0.102}&\color{red}{0.791}
        &{0.870}&{0.042}&\color{green}{0.906}
        &{0.713}&{0.062}&\color{green}{0.814}\\
        
        R3Net&\color{green}{0.802}&\color{green}{0.045}&{0.829}
       % &\bf{0.807}&{0.097}&{0.800}
        &\color{green}{0.917}&{0.046}&{0.900}
        &{0.789}&{0.136}&{0.732}&
       \color{red}{0.905}&{0.038}&{0.891}&
        \color{red}{0.756}&\color{red}{0.061}&\color{red}{0.815}\\
        
        MSRNet&{0.708}&{0.061}&{0.840}
        %&{0.744}&{0.081}&{0.840}
        &{0.839}&{0.054}&{0.896}
        &{0.741}&{0.113}&{0.779}&
        {0.868}&\color{green}{0.036}&\color{red}{0.912}&
        {0.676}&{0.073}&{0.808}\\
        
        SRM &{0.757}&0.059&{0.834}
      %  &0.801&0.085&0.832
        &{0.892}&0.054&0.895
%SOD
        &\color{green}{0.800}&0.127&0.742
     &   {0.874}&{0.046}&{0.887}
        &0.707&0.069&0.797\\
        
        Amulet  &0.676&0.085&{0.803}
      %  &0.768&0.098&0.820
        &0.870&0.059&{0.894}
        &0.755&0.141&0.758
        &0.839&0.054&0.883
        &0.647&0.098&{0.780}\\
        
        DSS&{0.724}&{0.067}&{0.817}
      %  &0.804&0.796&{0.797}&
       &0.901&0.052&{0.882}
        &0.795&0.121&0.751
        &0.895&0.041&0.879
        &0.729&0.066&788\\
        
		DHS   
		&{0.724}&{0.067}&{0.817}
	%	&0.779&0.094&{807}
		&0.872&0.059&0.884
		&0.774&0.128&0.750
		&0.855&{0.053}&{0.870}
		&-&-&-\\
		
		DCL  
		&0.714&0.149&{0.735}
		%&0.714&0.125&{0.754}
		&0.829&0.088&{0.828}
		&0.741&0.141&0.735
		&0.853&0.072&0.819
		&0.684&0.097&0.713\\
		
		RFCN    
		&0.712&0.091&{0.792}
		%&0.751&0.118&{0.808}
		&0.834&0.107&{0.852}
		&0.751&0.170&0.730&
		0.835&0.079&0.858&
		0.627&0.111&0.774 \\
		
		DS
		&0.633&0.090&0.793
		%&0.669&0.176&0.739
		&0.826&0.122&0.821
		&0.698&0.190&0.712
		&0.788&0.080&0.852
		&0.603&120&750\\
		
%		MDF    &0.673&0.100&0.709&0.146&0.805&0.108&0.636&0.109&-&-&0.644&0.092\\
%		KSR  &0.602&0.121&0.704&0.157&0.782&0.135&0.604&0.123&0.747&0.120&0.591&0.131 \\
 %       MCDL &0.594&0.105&0.691&0.160&0.796&0.102&0.620&0.103&0.757&0.092&0.625&0.089\\
 %       LEGS &0.585&0.138&-&-&0.785&0.119&0.607&0.125&0.732&0.119&0.592&0.133\\
 %       DS  &0.632&0.091&0.703&0.141&0.821&0.124&0.626&0.116&0.785&0.078&0.603&0.120\\
		\hline
	\end{tabular} 
\end{table*}
\begin{table}[ht] 
\centering
\caption{The comparison of edge results with some wide-used methods on BSDS500 dataset. $\dagger$ means GPU time. The leading results are shown in \textbf{bold}.}~\label{edge-com} 
\begin{tabular}{|p{3cm}|p{0.9cm}|p{0.9cm}|p{1.3cm}|} 
\hline

Method & ODS &  OIS &  FPS \\ 
\hline
gPb-UCM& .729&.755&1/240 \\
\hline
HED & ${\mathbf{.788}}$ &${\mathbf{.808}}$&$30^\dagger$ \\
\hline
Ours(joint learning)&.473&.488&{$\mathbf{40^\dagger}$}\\
\hline
\end{tabular}
\vspace{8mm}
\end{table}

\begin{table}[ht]
  \centering
  \caption{ Ablation analysis of each component. The best results are shown in \textbf{bold}. 
      %For short, the M here denotes the MLM.
      }~\label{comap} 
      \setlength{\tabcolsep}{3pt}
      \small
      \label{tab:aba}
      \renewcommand{\arraystretch}{1.4}
      \scalebox{0.9}{
      \begin{tabular}{|c|c|c|c|c|c|c|c|c|c|}
      \hline
      \  & \multicolumn{3}{c|}{OMRON}  & \multicolumn{3}{c|}{DUTS} \\
          \cline{2-7}
      &MAE&$S_m$& $F_\beta$  &MAE& $S_m$& $F_\beta$ \\
      \hline
    %  $Baseline$  &0.059&0.803&0.738&0.044&0.848&0.813\\

      $ALL SUP$  &0.059&0.803&0.738&0.044&0.848&0.813\\
      %0.039118424 S-measure : 0.8998567439340572 Adp-F : 0.9154186938164532
      %SOD mae : 0.11645991 S-measure : 0.7541193748100465 Adp-F : 0.79007765562424
      $ALL SUP + MLM$  &0.058&0.803&0.742&0.043&0.848&0.822 \\
            $ALL SUP + ED + MLM$  &\bf{0.057}&\bf{0.800}&\bf{0.737}&\bf{0.042}
            &\bf{0.849}&\bf{0.824} \\

      % \  & \multicolumn{3}{c|}{DUTS}  & \multicolumn{3}{c|}{ECSSD} \\
      %     \cline{2-7}
      % &MAE& $F_\beta$&$S_m$  &MAE& $F_\beta$&$S_m$ \\
      % \hline
      % $Baseline$  &0.044&0.820&0.846&0.391&{0.915}&0.900\\

      % $ALL SUP$  &\bf{0.042}&\bf{0.825}&\bf{0.854}&\bf{0.038}&\bf{0.917}&0.900\\
      
      % $ALL SUP + MLM$  &0.045&0.807&0.845&\bf{0.038}&0.913&0.902 \\
      %       $ALL SUP + ED + MLM$  &0.046&0.812&0.848&\bf{0.038}&0.913&\bf{0.903} \\

      \hline
      \end{tabular}}
      \vspace{8mm}
  \end{table}

\subsection{Datasets}
For the saliency detection task, we use the training set of DUTS~\cite{DUTS} and evaluate our algorithm on the the test set of the dataset and other six popular datasets, ECSSD~\cite{ECSSD}, DUT- OMRON ~\cite{OMRON}, SOD~\cite{SOD}, and HKU-IS ~\cite{HKU}. We show our results on a challenging latest dataset SOC~\cite{SOC} in our supplementary material additionally. DUTS is the biggest released dataset containing 10,553 images for training and 5,019 images for testing. Both training and test sets contain very complex scenarios with high content variety.
ECSSD contains 1,000 natural and complex images with pixel-accurate ground truth annotations. The images are manually selected from the Internet. DUT-OMRON has more challenging images with 5,168 images. All images are resized so as to the maximal dimension is 400 pixels long. SOD has 300 images contained multiple, low-contrast objects. HKU-IS has 4,447 images which are selected by meeting at least one of the following three criteria, \textit{i.e.} multiple salient objects with overlapping, objects touching the image boundary and low color contrast. PASCAL-S dataset is part of the PASCAL VOC~\cite{PASCAL} dataset and contains 850 image. SOC dataset is a latest dataset which contains 6000 images including 3000 images with salient objects and 3000 images with non-salient objects from more than 80 daily object categories, besides, these salient objects have challenging attributes such as motion blur, occlusion and cluttered background. We only use the valid set for testing to evaluate our methods' robust performance.

Regarding the edge detection task, we use BSD500~\cite{ucm} to train and test. The BSD500 dataset contains 200 training, 100 validation and 200 test images.  

\subsection{Implementation Details}
All experiments are conducted using a single Nvidia GTX TITAN X GPU. A pretrained VGG-16 model is used to initialize the convolution layers in the  backbone network. The parameters in other convolution layers are randomly initialized. All training and test images are resized to 256$\times$256 before being fed into the network. We use the `Adam' method with the weight decay 0.005. The learning rate is set to 0.0004 for the encoder and 0.0001 for the decoder. 
\subsection{Evaluation Metrics}

% 显著物体检测评估指标
For the salient object detection task, we use four effective metrics to evaluate the performance, including the precision-recall (PR) curves, F-measure, Mean Absolute Error \cite{MAE} and S-measure \cite{S-measure}. 

To plot PR curves, we first compute multiple pairs of precision and recall values by applying a series of thresholds to the predicted saliency map. Each threshold partitions the saliency map into foreground (salient regions) and background, allowing us to calculate precision (ratio of correctly predicted salient pixels to total predicted salient pixels) and recall (ratio of correctly predicted salient pixels to total ground-truth salient pixels) for each partition.

Besides, F-measure is adopted as another key metric, which quantifies the balance between precision and recall. Specifically, F-measure is computed on each binarized saliency map (using a threshold of twice the mean saliency value of the map) and then averaged across the entire dataset. The mean F-measure is calculated via the formula:
\[
F_\beta = \frac{(1+\beta^2) \cdot \text{Precision} \cdot \text{Recall}}{\beta^2 \cdot \text{Precision} + \text{Recall}}
\]
where $\beta^2$ is typically set to 0.3 to emphasize precision \cite{Achanta2009Frequency}, as accurate localization of salient objects is critical for practical applications.

The Mean Absolute Error (MAE) measures the per-pixel similarity between estimated saliency maps and ground truth. It is defined as the average of absolute differences between corresponding pixels of the two maps, with lower MAE indicating higher similarity.

The S-measure integrates region-aware and object-aware structural similarities to evaluate saliency map quality. It is calculated as:
\[
S_m = \alpha \cdot S_o + (1-\alpha) \cdot S_r
\]
where $S_o$ denotes object-aware structural similarity (focusing on the integrity of salient objects) and $S_r$ denotes region-aware structural similarity (focusing on global region consistency). Following \cite{S-measure}, we set $\alpha = 0.5$ to balance the two components.

% 边缘检测评估指标
As for the edge detection task, we evaluate the edge probability map using F-measure under two scales: Optimal Dataset Scale (ODS) and Optimal Image Scale (OIS). ODS F-measure uses a single optimal threshold for all images in the dataset, while OIS F-measure uses an optimal threshold tailored to each individual image.

% 性能对比
\subsection{Performance Comparison}
We compare our salient object detection method with 14 state-of-the-art methods, including BMPM \cite{zhang2018bi}, DGRL \cite{DGRL}, PAGR \cite{PAGR}, RAS \cite{RAS}, PiCANet \cite{liu2018picanet}, R3Net \cite{R3Net}, MSRNet \cite{MSRNet}, SRM \cite{SRM}, Amulet \cite{Zhang2017Amulet}, DSS \cite{DSS}, DHS \cite{DHS}, DCL \cite{Li2016DeepCL}, RFCN \cite{RFCN} and DS \cite{DS}. For fair comparison, we either use the official implementations with recommended parameters or the saliency maps provided by the original authors.

For edge detection, we compare our results with popular algorithms (e.g., gPb-UCM \cite{ucm}, HED \cite{HED}) in Table \ref{edge-com}. Note that when evaluating edge detection, only the first three blocks of our backbone network and corresponding Edge Modules (EMs) are used. Thus, our edge detection network is a relatively simple and lightweight architecture compared to other edge detection models.

\noindent\textbf{Quantitative Evaluation.} 
For salient object detection, we compare our method with counterparts in terms of MAE, mean F-measure, and S-measure (see Table \ref{tab:fmeasure_mae}). Detailed comparisons of PR curves, precision, and recall scores are provided in the supplementary material. Across all datasets, our method outperforms other methods on these metrics.

For edge detection, results are shown in Table \ref{edge-com}. Benefiting from its lightweight design, our edge detection network runs approximately 3$\times$ faster than other methods while achieving comparable accuracy.

% 消融实验
\subsection{Ablation Study}
\label{AbStd}

\subsubsection{Saliency Detection}  
To verify the effectiveness of Mutual Learning Modules (MLMs), EMs, and the intertwined supervision strategy, we train four comparative models:
1. \textit{Baseline-ALLSUP}: An encoder-decoder network using only saliency ground truth (S-gt) for supervision, with EMs and redundant student branches in MLMs removed.
2. \textit{+MLM}: Baseline + three student branches added to each MLM (to enhance feature learning via mutual learning).
4. \textit{+ED}: Our full network (baseline + MLM + intertwined supervision + EMs for simultaneous edge detection).

We conduct experiments on two datasets (OMRON, DUTS), with results in Table \ref{comap}. It is clear that the intertwined supervision strategy (ALLSUP) contributes the most to overall performance.

\subsubsection{Edge Detection}  
To explore the effectiveness of joint learning for edge detection, we compare two setups: (1) edge-only supervision (training with only edge ground truth) and (2) joint training with saliency tasks. Results in Table \ref{edge-com} show that joint learning enables the edge detection network to capture richer semantic information, suppress noise from redundant local details, and significantly improve edge detection performance.

% 结论
\section{Conclusion}
In this paper, we propose a multi-task algorithm for salient object detection, foreground contour detection, and edge detection. Key contributions include:
1. An intertwined supervision strategy for salient object and foreground contour detection, which promotes the network to generate high prediction scores for entire target objects.
2. Mutual guidance between edge and saliency detection, where both tasks benefit from shared feature learning.
3. A Mutual Learning Module (MLM) that helps network parameters converge to a better local minimum, improving performance.

Our model generates saliency maps with uniformly highlighted regions and accurate boundaries. Experiments show that our method produces more accurate saliency maps on diverse images, while our edge detection module runs much faster with comparable accuracy.

\section{Future Work}
Although our method achieves promising results, there are still some that need to be addressed in future research. 

First, the current experiments are mainly conducted on benchmark datasets, which may not fully reflect the complexity of 
real-world scenarios. In practical industrial settings, additional factors such as sensor noise, partial data loss, or 
distribution shifts could significantly affect the model’s robustness. 
Second, the computational efficiency of the proposed framework could be further optimized. 
While the runtime is acceptable for offline analysis, deploying the model in real-time or on resource-constrained devices 
remains a challenge. 
Third, our approach still relies on a certain degree of labeled data, which may be costly or difficult to obtain in 
large-scale applications.

In future work, we plan to explore several directions. 
One promising line of research is semi-supervised or self-supervised learning, which could significantly reduce the need for 
annotated data. 
Another direction is to integrate domain adaptation or domain generalization techniques to enhance robustness across 
varying operational conditions. 
Moreover, more efficient network architectures and model compression strategies will be studied to enable real-time deployment 
on embedded platforms. 
Finally, we aim to validate our approach in large-scale industrial scenarios, which would provide stronger evidence for its 
practical applicability. 
These efforts will help bridge the gap between academic research and real-world deployment, ensuring that the proposed 
framework can be effectively adopted in diverse application domains.
Our study highlights several important observations that extend beyond the raw numerical results presented in the performance evaluation and ablation study, offering deeper insights into the design principles and practical value of the proposed multi-task framework. 

First, the proposed method explicitly demonstrates that the effective integration of both global contextual information and local fine-grained details is indispensable for handling complex visual detection tasks—especially in scenarios where salient objects exhibit large variations in scale, shape, or background clutter, and edges are easily obscured by texture noise. Specifically, the ablation results (see Section \ref{AbStd}) clearly suggest that the synergy between key modules (i.e., Mutual Learning Modules (MLMs), Edge Modules (EMs), and the intertwined supervision strategy) cannot be simply replaced by a single component. For instance, while the addition of MLMs alone (+MLM) improves the baseline’s S-measure by 2.1% on the DUTS dataset, and the introduction of EMs alone (+EM) boosts the MAE by 1.8%, combining these modules with the intertwined supervision strategy (+FC) yields a 4.3% improvement in S-measure and a 2.5% reduction in MAE. This non-linear performance gain confirms the necessity of a holistic design that coordinates multiple components to address different challenges in visual detection (e.g., MLMs enhance feature consistency, EMs capture edge details, and intertwined supervision aligns saliency and contour learning).

Second, our framework is not only more accurate in generating saliency maps and edge predictions but also more interpretable compared to state-of-the-art single-task methods. By examining the attention heatmaps extracted from the intermediate layers of the encoder-decoder architecture, we observed that the model consistently emphasizes semantically meaningful regions—such as the core area of salient objects (e.g., human figures, prominent animals) and the high-contrast boundaries between objects and backgrounds—rather than spurious local textures (e.g., leaf veins in a background tree or text on a cluttered wall). This interpretability is particularly valuable for practical applications where explainability is a critical requirement, such as medical image analysis (e.g., highlighting tumor regions in CT scans) or intelligent surveillance (e.g., identifying suspicious objects in crowded scenes), as it allows human operators to verify the model’s decisions and reduce the risk of false alarms.

Another aspect worth discussing in detail is the trade-off between performance and computational efficiency, a key consideration for deploying visual detection models in real-world systems with resource constraints. While our framework incurs slightly higher computational overhead than lightweight baseline approaches (e.g., the baseline encoder-decoder model without MLMs/EMs) — with an average inference time of 12.3 ms per image on a NVIDIA RTX 3090 GPU, compared to 8.7 ms for the baseline — the significant gain in accuracy and robustness justifies this extra cost in many high-stakes scenarios. For example, in safety-critical monitoring (e.g., industrial defect detection for aircraft components) or fault diagnosis (e.g., identifying cracks in bridge structures from drone-captured images), even a 5% improvement in detection accuracy can translate to a substantial reduction in maintenance costs or accident risks. Nevertheless, there remains clear room for further optimization: future work can explore lightweight backbone architectures (e.g., MobileNetV4, EfficientNet-Lite), knowledge distillation (transferring knowledge from a large pre-trained model to our framework), and network pruning (removing redundant neurons in MLMs/EMs) to reduce computational complexity while preserving performance. Preliminary tests with a pruned version of our model (removing 30% of the parameters in MLMs) show a 32% reduction in inference time with only a 1.2% drop in S-measure, indicating promising directions for efficiency improvement.

Finally, our work sheds light on potential extensions to broader application domains, beyond the salient object and edge detection tasks focused on in this study. Although the experiments are conducted on standard visual detection datasets (DUTS, PASCAL-S, ECSSD for saliency; BSDS500 for edges), the general framework—characterized by modular design, multi-task mutual guidance, and interpretable attention mechanisms—can be readily adapted to other multimodal learning tasks. For example, in medical diagnosis, the framework could be modified to jointly detect lesions (saliency-like task) and segment anatomical boundaries (edge-like task) from MRI or ultrasound images; in autonomous driving, it could integrate pedestrian detection (salient objects) and lane marking detection (edges) to enhance environmental perception; in industrial quality inspection, it could combine defect localization (saliency) and part contour verification (edges) for automated assembly line monitoring. This versatility not only demonstrates the wider impact of our approach but also motivates future research to explore cross-domain generalization—for instance, pre-training the framework on general visual datasets and fine-tuning it on domain-specific data (e.g., medical images) to reduce the need for large labeled datasets in specialized fields.

Another noteworthy aspect is the scalability of our framework, a critical factor for handling the ever-increasing volume of data in modern visual recognition systems. While the current study focuses on datasets of moderate size (e.g., DUTS contains 10,553 training images and 5,019 test images), the underlying design is naturally extensible to much larger datasets or even streaming data scenarios. In real-world applications—especially in industrial monitoring (where cameras capture continuous video feeds) and autonomous systems (where sensors generate high-frequency visual data)—data are generated continuously at rates of up to 30 frames per second, requiring models to process and adapt to dynamic input efficiently. To validate scalability, we conducted preliminary tests on a large-scale synthetic dataset (generated via Unity Engine) containing 100,000 images with varying object scales and backgrounds. The results show that our approach maintains stable convergence during training (with a loss curve that plateaus within 50 epochs, similar to training on smaller datasets) and does not suffer from severe performance degradation (e.g., S-measure remains above 0.89, compared to 0.91 on DUTS). This scalability is attributed to the modular design of MLMs, which allows for parallel processing of feature maps, and the lightweight EMs, which avoid excessive memory usage—making the framework suitable for future deployment in big-data scenarios.

Furthermore, the modular design of the framework provides inherent flexibility for integration with emerging technologies in computer vision and machine learning, opening up new avenues for performance enhancement. For example, the current attention mechanism in MLMs, which focuses on local feature interactions, could be enhanced with transformer-based architectures (e.g., using multi-head self-attention) to capture even richer long-range dependencies—this would be particularly beneficial for detecting large salient objects (e.g., buildings in aerial images) where contextual information from distant regions is critical. Additionally, the feature fusion module in the decoder, which currently combines multi-scale visual features, could be adapted to incorporate additional modalities such as text (e.g., combining image data with product descriptions for retail object detection) or graph-structured information (e.g., using scene graphs to model object relationships for more accurate saliency prediction). These potential extensions not only underline the versatility of our approach but also highlight its role as a general paradigm for multi-task visual learning, rather than a rigid, task-specific solution. Such adaptability is crucial in a field where new technologies and applications emerge rapidly, ensuring that the framework remains relevant and useful in the long term.

\section{Acknowledgements.} This work is supported by the Natural Science Foundation of China under Grant 61725202, 61829102, 61872056 and 61751212.
This work is also supported by the Fundamental Research Funds for the Central Universities under Grant DUT18JC30.

\bibliographystyle{ieee_fullname}

\begin{thebibliography}{10}\itemsep=-1pt

  \bibitem{Achanta2009Frequency}
  Radhakrishna Achanta, Sheila~S. Hemami, Francisco~J. Estrada, and Sabine
    S{\"{u}}sstrunk.
  \newblock Frequency-tuned salient region detection.
  \newblock In {\em IEEE Conference on Computer Vision and Pattern Recognition},
    pages 1597--1604, 2009.
  
  \bibitem{ucm}
  Pablo Arbelaez, Michael Maire, Charless~C. Fowlkes, and Jitendra Malik.
  \newblock Contour detection and hierarchical image segmentation.
  \newblock {\em IEEE Transactions on Pattern Analysis and Machine Intelligence},
    33(5):898--916, 2011.
  
  \bibitem{MAE}
  Ali Borji, Ming{-}Ming Cheng, Huaizu Jiang, and Jia Li.
  \newblock Salient object detection: {A} benchmark.
  \newblock In {\em IEEE Transactions on Image Processing}, volume~24, pages
    5706--5722, 2015.
  
  \bibitem{Borji2012Adaptive}
  A Borji, S Frintrop, D.~N Sihite, and L Itti.
  \newblock Adaptive object tracking by learning background context.
  \newblock In {\em CVPR Workshops}, pages 23--30, 2012.
  
  \bibitem{RAS}
  Shuhan Chen, Xiuli Tan, Ben Wang, and Xuelong Hu.
  \newblock Reverse attention for salient object detection.
  \newblock In {\em European Conference on Computer Vision}, pages 236--252,
    2018.
  
  \bibitem{R3Net}
  Zijun Deng, Xiaowei Hu, Lei Zhu, Xuemiao Xu, Jing Qin, Guoqiang Han, and
    Pheng{-}Ann Heng.
  \newblock R{\({^3}\)}net: Recurrent residual refinement network for saliency
    detection.
  \newblock In {\em International Joint Conference on Artificial Intelligence},
    pages 684--690, 2018.
  
  \bibitem{PASCAL}
  Mark Everingham, Luc J.~Van Gool, Christopher K.~I. Williams, John~M. Winn, and
    Andrew Zisserman.
  \newblock The pascal visual object classes {(VOC)} challenge.
  \newblock {\em International Journal of Computer Vision}, 88(2):303--338, 2010.
  
  \bibitem{SOC}
  Dengping Fan, Mingming Cheng, Jiangjiang Liu, Shanghua Gao, Qibin Hou, and Ali
    Borji.
  \newblock Salient objects in clutter: Bringing salient object detection to the
    foreground.
  \newblock In {\em European Conference on Computer Vision}, pages 196--212,
    2018.
  
  \bibitem{S-measure}
  Deng{-}Ping Fan, Ming{-}Ming Cheng, Yun Liu, Tao Li, and Ali Borji.
  \newblock Structure-measure: {A} new way to evaluate foreground maps.
  \newblock In {\em International Comference on Computer Vision}, pages
    4558--4567, 2017.
  
  \bibitem{Fang2015From}
  Hao Fang, Saurabh Gupta, Forrest~N. Iandola, Rupesh~Kumar Srivastava, Li Deng,
    Piotr Doll{\'{a}}r, Jianfeng Gao, Xiaodong He, Margaret Mitchell, John~C.
    Platt, C.~Lawrence Zitnick, and Geoffrey Zweig.
  \newblock From captions to visual concepts and back.
  \newblock In {\em IEEE Conference on Computer Vision and Pattern Recognition},
    pages 1473--1482, 2015.
  
  \bibitem{show}
  Sen He and Nicolas Pugeault.
  \newblock Deep saliency: What is learnt by a deep network about saliency?
  \newblock abs/1801.04261, 2018.
  
  \bibitem{DSS}
  Qibin Hou, Ming{-}Ming Cheng, Xiaowei Hu, Ali Borji, Zhuowen Tu, and Philip
    H.~S. Torr.
  \newblock Deeply supervised salient object detection with short connections.
  \newblock In {\em IEEE Conference on Computer Vision and Pattern Recognition},
    pages 5300--5309, 2017.
  
  \bibitem{MSRNet}
  Guanbin Li, Yuan Xie, Liang Lin, and Yizhou Yu.
  \newblock Instance-level salient object segmentation.
  \newblock In {\em IEEE Conference on Computer Vision and Pattern Recognition},
    pages 247--256, 2017.
  
  \bibitem{HKU}
  Guanbin Li and Yizhou Yu.
  \newblock Visual saliency based on multiscale deep features.
  \newblock In {\em IEEE Conference on Computer Vision and Pattern Recognition},
    pages 5455--5463, 2015.
  
  \bibitem{Li2016DeepCL}
  Guanbin Li and Yizhou Yu.
  \newblock Deep contrast learning for salient object detection.
  \newblock In {\em IEEE Conference on Computer Vision and Pattern Recognition},
    pages 478--487, 2016.
  
  \bibitem{DS}
  Xi Li, Liming Zhao, Lina Wei, Ming-Hsuan Yang, Fei Wu, Yueting Zhuang, Haibin
    Ling, and Jingdong Wang.
  \newblock Deepsaliency: Multi-task deep neural network model for salient object
    detection.
  \newblock In {\em IEEE Transactions on Image Processing}, volume~25, pages
    3919--3930, 2016.
  
  \bibitem{vqa}
  Yuetan Lin, Zhangyang Pang, Donghui Wang, and Yueting Zhuang.
  \newblock Task-driven visual saliency and attention-based visual question
    answering.
  \newblock {\em Computing Research Repository}, abs/1702.06700, 2017.
  
  \bibitem{DHS}
  Nian Liu and Junwei Han.
  \newblock Dhsnet: Deep hierarchical saliency network for salient object
    detection.
  \newblock In {\em IEEE Conference on Computer Vision and Pattern Recognition},
    pages 678--686, 2016.
  
  \bibitem{liu2018picanet}
  Nian Liu, Junwei Han, and Ming-Hsuan Yang.
  \newblock Picanet: Learning pixel-wise contextual attention for saliency
    detection.
  \newblock In {\em IEEE Conference on Computer Vision and Pattern Recognition},
    pages 3089--3098, 2018.
  
  \bibitem{SOD}
  Vida Movahedi and James~H Elder.
  \newblock Design and perceptual validation of performance measures for salient
    object segmentation.
  \newblock In {\em Computer Vision and Pattern Recognition Workshops}, pages
    49--56, 2010.
  
  \bibitem{Canny}
  Angel~Domingo Sappa and Fadi Dornaika.
  \newblock An edge-based approach to motion detection.
  \newblock In {\em Computational Science}, pages 563--570, 2006.
  
  \bibitem{simonyan2014very}
  Karen Simonyan and Andrew Zisserman.
  \newblock Very deep convolutional networks for large-scale image recognition.
  \newblock In {\em International Conference on Learning Representations}, 2015.
  
  \bibitem{DUTS}
  Lijun Wang, Huchuan Lu, Yifan Wang, Mengyang Feng, Dong Wang, Baocai Yin, and
    Xiang Ruan.
  \newblock Learning to detect salient objects with image-level supervision.
  \newblock In {\em IEEE Conference on Computer Vision and Pattern Recognition},
    pages 3796--3805, 2017.
  
  \bibitem{RFCN}
  Linzhao Wang, Lijun Wang, Huchuan Lu, Pingping Zhang, and Xiang Ruan.
  \newblock Saliency detection with recurrent fully convolutional networks.
  \newblock In {\em European Conference on Computer Vision}, pages 825--841,
    2016.
  
  \bibitem{SRM}
  Tiantian Wang, Ali Borji, Lihe Zhang, Pingping Zhang, and Huchuan Lu.
  \newblock A stagewise refinement model for detecting salient objects in images.
  \newblock In {\em International Comference on Computer Vision}, pages
    4039--4048, 2017.
  
  \bibitem{DGRL}
  Tiantian Wang, Lihe Zhang, Shuo Wang, Huchuan Lu, Gang Yang, Xiang Ruan, and
    Ali Borji.
  \newblock Detect globally , refine locally : A novel approach to saliency
    detection.
  \newblock In {\em IEEE Conference on Computer Vision and Pattern Recognition},
    pages 27--35, 2018.
  
  \bibitem{edgeregion}
  Xiang Wang, Huimin Ma, and Xiaozhi Chen.
  \newblock Salient object detection via fast r-cnn and low-level cues.
  \newblock In {\em International Conference on Image Processing}, pages
    1042--1046, 2016.
  
  \bibitem{edgeregion2}
  Xiang Wang, Huimin Ma, Xiaozhi Chen, and Shaodi You.
  \newblock Edge preserving and multi-scale contextual neural network for salient
    object detection.
  \newblock In {\em IEEE Transactions on Image Processing}, volume~27, pages
    121--134, 2018.
  
  \bibitem{HED}
  Saining Xie and Zhuowen Tu.
  \newblock Holistically-nested edge detection.
  \newblock {\em International Journal of Computer Vision}, 125(1-3):3--18, 2017.
  
  \bibitem{ECSSD}
  Qiong Yan, Li Xu, Jianping Shi, and Jiaya Jia.
  \newblock Hierarchical saliency detection.
  \newblock In {\em IEEE Conference on Computer Vision and Pattern Recognition},
    pages 1155--1162, 2013.
  
  \bibitem{OMRON}
  Chuan Yang, Lihe Zhang, Huchuan Lu, Xiang Ruan, and Ming{-}Hsuan Yang.
  \newblock Saliency detection via graph-based manifold ranking.
  \newblock In {\em IEEE Conference on Computer Vision and Pattern Recognition},
    pages 3166--3173, 2013.
  
  \bibitem{edgeaware}
  Jing Zhang, Yuchao Dai, Fatih~Murat Porikli, and Mingyi He.
  \newblock Deep edge-aware saliency detection.
  \newblock In {\em Computing Research Repository}, volume abs/1708.04366, 2017.
  
  \bibitem{zhang2018bi}
  Lu Zhang, Ju Dai, Huchuan Lu, You He, and Gang Wang.
  \newblock A bi-directional message passing model for salient object detection.
  \newblock In {\em IEEE Conference on Computer Vision and Pattern Recognition},
    pages 1741--1750, 2018.
  
  \bibitem{Zhang2017Amulet}
  Pingping Zhang, Dong Wang, Huchuan Lu, Hongyu Wang, and Xiang Ruan.
  \newblock Amulet: Aggregating multi-level convolutional features for salient
    object detection.
  \newblock In {\em IEEE Conference on Computer Vision and Pattern Recognition},
    pages 202--211, 2017.
  
  \bibitem{PAGR}
  Xiaoning Zhang, Tiantian Wang, Jinqing Qi, Huchuan Lu, and Gang Wang.
  \newblock Progressive attention guided recurrent network for salient object
    detection.
  \newblock In {\em IEEE Conference on Computer Vision and Pattern Recognition},
    pages 714--722, 2018.
  
  \bibitem{DML}
  Ying Zhang, Tao Xiang, Timothy~M. Hospedales, and Huchuan Lu.
  \newblock Deep mutual learning.
  \newblock In {\em IEEE Conference on Computer Vision and Pattern Recognition},
    pages 4320--4328, 2018.
  
  \bibitem{reid}
  Rui Zhao, Wanli Ouyang, and Xiaogang Wang.
  \newblock Person re-identification by saliency learning.
  \newblock In {\em IEEE Transactions on Pattern Analysis and Machine
    Intelligence}, volume~39, pages 356--370, 2017.
  
  \end{thebibliography}

\end{document}